\documentclass[conference]{IEEEtran}
\IEEEoverridecommandlockouts
\usepackage{cite}
\usepackage{amsmath,amssymb,amsfonts}
\usepackage{algorithmic}
\usepackage{graphicx}
\usepackage{textcomp}
\usepackage{listings}
\usepackage{xcolor}
\def\BibTeX{{\rm B\kern-.05em{\sc i\kern-.025em b}\kern-.08em
    T\kern-.1667em\lower.7ex\hbox{E}\kern-.125emX}}
\begin{document}

\title{Safe, Untrusted, ``Proof-Carrying'' AI Agents: \\toward the agentic lakehouse\\
\thanks{Thanks to \cite{10.5555/648051.746192} for coming up with a great title (a long time ago, for a different type of agents).}
}

\author{\IEEEauthorblockN{Jacopo Tagliabue}
\IEEEauthorblockA{\textit{Bauplan Labs} \\
NYC, USA \\
jacopo.tagliabue@bauplanlabs.com}
\and
\IEEEauthorblockN{Ciro Greco}
\IEEEauthorblockA{\textit{Bauplan Labs} \\
NYC, USA \\
ciro.greco@bauplanlabs.com}
}

\maketitle

\begin{abstract}
Data lakehouses run sensitive workloads, where AI-driven automation raises concerns about trust, correctness, and governance. We argue that API-first, programmable lakehouses provide the right abstractions for \textit{safe-by-design}, agentic workflows. Using Bauplan as a case study, we show how data branching and declarative environments extend naturally to agents, enabling reproducibility and observability while reducing the attack surface. We present a proof-of-concept in which agents repair data pipelines using correctness checks inspired by proof-carrying code. Our prototype demonstrates that untrusted AI agents can operate safely on production data and outlines a path toward a fully agentic lakehouse.
\end{abstract}



\section{Introduction}

The data lakehouse is the \textit{de facto} cloud architecture for analytics and Artificial Intelligence (AI) workloads \cite{mazumdar2023datalakehousedatawarehousing,Zaharia2021LakehouseAN}, thanks to storage-compute decoupling, multi-language support, and unified table semantics. As reasoning and tool usage in Large Language Models (LLMs) improve \cite{shen2024llm}, autonomous decisions (``AI agents'') are both supported by, and targeted at, cloud infrastructure: to what extent can agents manage the data lifecycle in a lakehouse?

\textit{Prima facie}, the question appears both too hard and too broad. On one hand, lakehouses are distributed systems built for the collaboration of human teams on sensitive production data, not point-wise tasks immediately suitable for end-to-end automation. On the other, it is unclear how to prioritize agentic use cases across such heterogeneous platforms. \textit{This} paper is a preliminary answer to these challenges: we detail lakehouse abstractions suitable for automation, and operationalize a prototype for an important use case: \textit{repairing data pipelines}.

Pipelines are a compelling case study for three reasons: first, they cover a large portion of lakehouse workloads, measured both by developer time \cite{cdms2025eudoxia} and overall compute \cite{Renen2024}. Second, data engineers spend a significant amount of their time fixing them \cite{dataworld,wang2022empirical}. Finally, repairing pipelines is a canary test for agent penetration in high-stakes non-trivial scenarios, which are often hard even for expert humans \cite{Yasmin2024AnES,FOIDL2024111855}. We summarize our contributions as follows:

\begin{enumerate}
     \item we introduce abstractions to model the data life-cycle in a programmable lakehouse \cite{10.1145/3702634.3702955}, i.e. building and executing cloud pipelines entirely through \textit{code}. We argue that traditional systems resist automation mostly because of heterogeneous interfaces and complex access patterns, while code is the \textit{lingua franca} suitable for agents, cloud systems, and human supervisors;   
     \item we review common objections to automating high-stakes workloads in light of the proposed abstractions: in particular, we argue that our model promotes trustworthiness and correctness both in data and code artifacts;
     \item we release working code\footnote{Open source code is  available at https://github.com/BauplanLabs/the-agentic-lakehouse.}, showing a proof of concept for self-repairing pipelines using \texttt{Bauplan} as a lakehouse and an agentic loop. Starting from this prototype, we conclude by outlining practical next steps for a full agentic lakehouse.
\end{enumerate}

The paper is organized as follows. After reviewing agent-friendly abstractions (Section~\ref{sec:background}), we address key safety objections for high-stakes scenarios (Section~\ref{sec:safety}). Once safety is established, we describe a ReAct \cite{yao2023reactsynergizingreasoningacting}  loop built on these abstractions (Section~\ref{sec:poc}). We put forward our working prototype as a feasibility demonstration of safe-by-design data agents, not as a full-fledged experimental benchmark.

We believe that sharing working code is of great value to the community, especially in times of quickly shifting mental models. However, it is important to remember that our fundamental insights -- programmability and safety -- can be replicated independently of the chosen APIs. For these reasons, we believe our paper to be valuable to a wide range of practitioners: on one hand, those looking for a new mental map of this uncharted territory; on the other, those looking to be inspired by tinkering with existing implementations and inspecting systems working at scale.

\section{A programmable lakehouse}
\label{sec:background}

In a \textit{programmable} lakehouse, the \textit{entire} data life-cycle -- data, user and infrastructure management, pipeline and query execution, runtime observability -- is exposed through code abstractions: server-side APIs, SDK methods, CLI shortcuts. In the rest of the paper, \texttt{Bauplan} snippets will be used as sample implementation, but the platform’s composable nature makes it easy to replicate these functionalities in different architectures.\footnote{We refer the interested reader to existing papers, in particular  \cite{10.1145/3702634.3702955,10825377,10.1145/3650203.3663335}.} We break down the pipeline life-cycle into two major components: pipeline definition and pipeline execution.

\subsection{Pipeline definition}
\label{sec:def}

\begin{figure*}
    \centering
    \includegraphics[width=\linewidth]{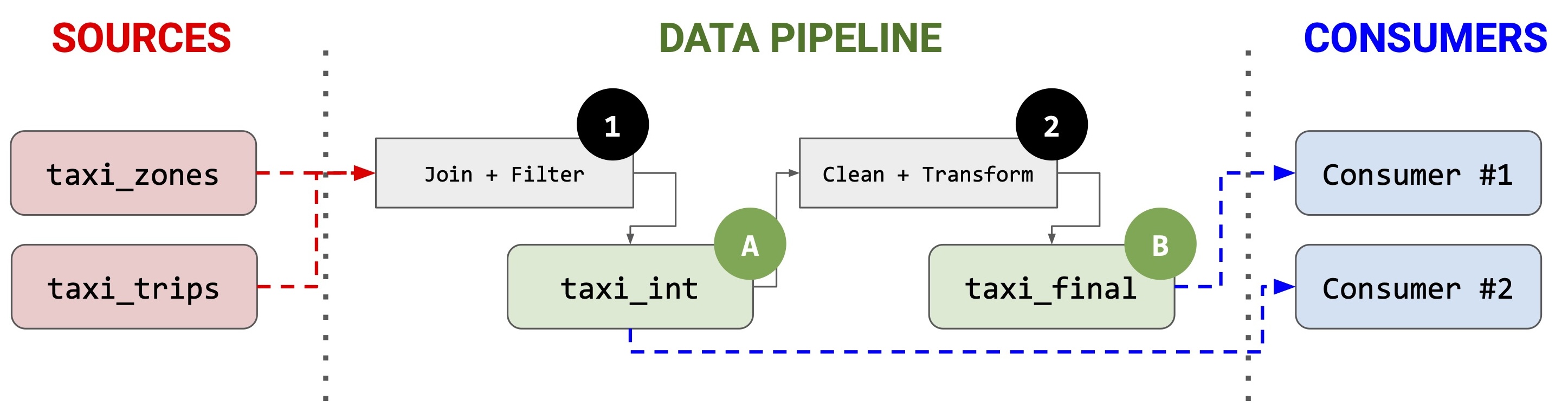}
    \caption{\textbf{Sample pipeline}: upstream from the pipeline, two source tables containing taxi trips and location data, downstream, multiple consumers. The pipeline itself is a two-node DAG, with compute steps in \textit{gray} -- 1 and 2 --, and tables in \textit{green} -- A and B.}
    \label{fig:dag}
\end{figure*}

A pipeline is a DAG of transformations. A DAG starts from source tables, which are progressively cleaned, augmented, aggregated through the transformation code (expressed in SQL or Python). A successful execution produces intermediate and final data assets, which are then consumed by downstream systems \cite{salami2025hubstarmodeling20}. Fig.~\ref{fig:dag} shows a pipeline (hence, P) used as a recurring example throughout the paper. Taking two tables from the NYC taxi dataset \cite{nyctaxi}, P defines two new tables (A and B), based on two transformations (1 and 2). The Bauplan implementation for P is as follows:\footnote{Snippets are simplified in the interest of space: full code is available in the open source repository.}

\addvspace{\baselineskip}
\begin{lstlisting}[showstringspaces=false,columns=fullflexible,language=Python,caption=P as the Python file \textit{p.py}]

@bauplan.model(materialization="REPLACE", name="A")
@bauplan.python("3.10", pip={"pandas": "2.0"})
def join_and_filter(
    trips=bauplan.Model("taxi_trips"),
    zones=bauplan.Model("taxi_zones") 
):
   # some transformation here...
   return trips.join(zones).do_something()

@bauplan.model(materialization="REPLACE", name="B")
@bauplan.python("3.11", pip={"pandas": "1.5.3"})
def clean_and_transform(
   data=bauplan.Model("join_and_filter") 
):
   return data.do_something()
\end{lstlisting}
\addvspace{\baselineskip}

Two important design choices are worth highlighting in connection to our safety discussion (Section~\ref{sec:safety}):

\begin{itemize}
    \item \textbf{Function-as-a-service (FaaS) abstractions}: business logic is expressed in the body of plain vanilla functions with the signature $Table(s) \rightarrow Table$. DAGs are functions chained through naming convention. These abstractions naturally map to a serverless runtime, which can execute the requested computation efficiently \cite{10.1145/3702634.3702955};
    \item \textbf{declarative I/O and infrastructure}: functions are fully isolated (e.g. two functions, two versions of \texttt{pandas}) and their Python environment is specified declaratively  \cite{tagliabue2023reasonablescalemachinelearning}. Reading tables and writing artifacts back to the lake is also fully declarative: users specify the needed inputs and desired output, the platform performs the corresponding physical operations.
\end{itemize}

\subsection{Pipeline execution}
\label{sec:exec}

A human (or an agent) with the proper access can execute \texttt{p.py} by simply installing the \texttt{bauplan} package, and running it from the terminal, without any additional steps -- no Docker, no Terraform, no JDBC clients:\footnote{To get a first-person perspective on the developer experience, the reader is invited to pause and watch a recorded run before continuing: https://www.loom.com/share/99ac0d5b5f944fc9aeef132bfaea0881}

\addvspace{\baselineskip}
\begin{lstlisting}[showstringspaces=false,columns=fullflexible,language=bash,caption=Bauplan CLI]
$ pip install bauplan
$ bauplan run --project_dir P_folder
\end{lstlisting}
\addvspace{\baselineskip}

\begin{figure}
    \centering
    \includegraphics[width=\linewidth]{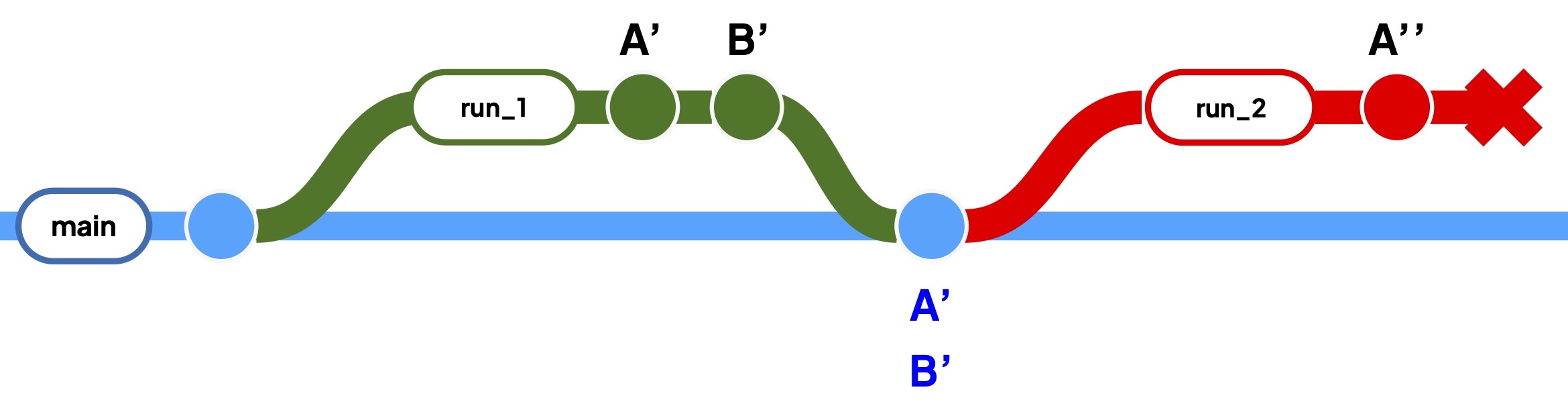}
    \caption{\textbf{Transactional pipelines}: a successful execution of P, followed by one that failed after A materialization. Runs happen on data branches: input data is from sources in \texttt{main}, but writes are sandboxed so that materialized tables hit \texttt{main} \textit{atomically} only on success (no half-written pipeline).}
    \label{fig:transaction}
\end{figure}

While simplicity is a virtue, for our present concerns we focus on the \textit{transactional} nature of the \texttt{run} API, which is obtained by re-purposing Git concepts to table evolution, and by managing data and compute as a logical whole even if physically decoupled. Consider now the two runs depicted in Fig.~\ref{fig:transaction}: \texttt{run 1} (successful) and \texttt{run 2} (failed). The branches and merges in the picture are a graphical representation of the underlying ``Git-for-Data'' abstractions \cite{Tagliabue2023BuildingAS}: if every change to the lake corresponds to a \textit{commit}, a \textit{branch} is the HEAD of a sequence of commits, and a \textit{merge} atomically combines commits from two branches. If the \texttt{main} branch represents production, \textit{merge} operations between on-branch writes and \textit{main} mimic software best practices, and provides a natural hook for testing and reviews.

When \texttt{run 1} starts, the execution is automatically moved to a copy-on-write branch: source tables will have the same data as production, but the tables materialized by the run will be written to this branch first and then merged to \texttt{main} with an atomic operation, promoting A' and B' as the new tables for downstream consumers. The importance of coupling runs with a \textit{branch-then-merge} pattern becomes evident with \texttt{run 2}, which failed before a new version of B was materialized: no merge happens for \texttt{run 2}, so no dirty read can occur in \texttt{main} -- downstream systems will still read a consistent pair, A' and B', not A'' and B'. In other words, \textit{branches} allow sand-boxed writes starting from production reads, i.e. working with production data without the risk of destroying production. 

Git-for-Data abstractions at the execution level therefore complement the functional abstractions at the definition level, solving three important use cases through simple APIs:

\begin{itemize}
    \item \textbf{reproducibility}: runs are immutably, deterministically identified through a pointer to the starting \textit{commit} and a copy of the code;
    \item \textbf{transactionality}: the \textit{branch-then-merge} pattern allows transactional pipelines, i.e. doing MVCC in a ``deconstructed database'' \cite{cerone_et_al:LIPIcs.CONCUR.2015.58,under};
    \item \textbf{reversibility}: writes are immutable but never final, as we can always revert to a previous state through the relevant commit.
\end{itemize}

\subsection{Code as the universal interface}

Agents are a combination of reasoning (provided by LLMs) and tool usage. In a lakehouse, tools should allow \textit{observability} of present and past runs and \textit{reproducibility} of existing pipelines: if a lakehouse offers these capabilities in typed, documented methods, exposing them for agentic usage is as trivial as wrapping those methods in MCP routes. In other words, from the point of view of the relevant abstractions -- functions, decorators, data branches --, a programmable lakehouse \textit{is already} an agentic lakehouse.

In this sense, while serving overlapping use cases, the design distance between a programmable lakehouse and traditional systems could not be greater. Practitioners today have to context-switch between one-off Spark clusters, development notebooks, SQL editors, cloud warehouses and more \cite{cdms2025eudoxia}; \textit{code}, on the other hand, is a universally understood interface -- easy for agents to master, immediate for the cloud to run \textit{and} safe for humans to properly supervise. 

\section{The Safety Checklist}
\label{sec:safety}
Making agents able to ``code the lakehouse'' through purpose-built abstractions is a necessary, but not sufficient condition to repair pipelines in production. We also need to make sure that the proposed abstractions address \textit{trust and correctness} concerns over malicious or (simply wrong) usage (Table~\ref{safety_table}).

\textbf{Trust in data}: can agents access data they are not supposed to? No. I/O is always mediated by the platform: agents have no access to the physical data layer (S3), so reads and writes happen always in platform space, not user space. More generally, since \textit{all} operations are API operations, RBAC over API keys provides fine-grained permissions with a minimal attack surface.

\textbf{Trust in code}: can agents run untrusted packages or access malicious web resources? No. Functions are run as independent processes isolated from their host and other functions, with \textit{no internet access} (since I/O is in platform space, users do not even access S3!). The declarative syntax makes (dis)allowing packages as trivial as checking a decorator against a whitelist. Once again, a concise surface comes in handy, as there is only one entry point for dependency management.

\textbf{Correctness in data}: can agents damage production data? No. First, incomplete pipelines will not affect downstream systems (Section~\ref{sec:exec}); second, removing merge-to-main permissions will make a human  review necessary to reach production; third, humans can \textit{always} revert tables using past commits.

\textbf{Correctness in code}: can agents push to production silent bugs? No (as in, no more than humans can). The key insight is that even untrusted code can be useful provided a hard-to-fake correctness test. The aptly titled \cite{10.5555/648051.746192} defines an interesting protocol: a ``consumer'' specifies safety conditions upfront, a ``producer'' provides evidence that its work satisfies them. If the consumer is satisfied, the work is now ``trusted''. We can imagine a similar protocol, where the ``consumer'' is the lakehouse owner (i.e. a data engineer), and the ``producer'' is the agent repairing the pipeline on her behalf. The evidence will revolve around the pipeline output (A and B in P) meeting certain criteria. In the prototype below, the owner came up with a verifier, i.e. a function $Branch \rightarrow bool$ to allow an agent branch to be merged. Unlike the original protocol, data verifiers are less about formal properties and more about the business context for the generated tables: the same principles apply though, here strengthened by the ``pull-request'' flow that Git-for-Data enables. Importantly, verifiers use the same APIs as the agents, leveraging once again the unified access to data and compute, with no semantic or infrastructure drift between lakehouse clients.

\begin{table}[!t]
\renewcommand{\arraystretch}{1.3}
\caption{Mapping safety concerns to abstractions}
\label{safety_table}
\centering
\begin{tabular}{c|c|c}
\hline
\bfseries Concern & \bfseries Mode & \bfseries Abstraction \\
\hline
Trust & Data & Declarative I/O\\
\hline
Trust & Code & FaaS runtime\\
\hline
Correctness & Data & Transactional runs\\
\hline
Correctness & Code & Verify-then-merge\\
\hline
\end{tabular}
\end{table}

\section{A Proof of Concept}
\label{sec:poc}

\texttt{run 2} failed (Fig.~\ref{fig:transaction}): can an agent repair it? Now that we know it is safe to do so, we combine what we learned so far into an agentic loop and draw some preliminary conclusions from the prototype. 

\subsection{The self-repairing setup}

Our setup is as follows:

\begin{itemize}
    \item Bauplan as the programmable lakehouse;
    \item the Bauplan MCP\footnote{https://github.com/BauplanLabs/bauplan-mcp-server}, exposing as tools the lakehouse APIs: tools can be used for \textit{observability} (e.g. get failed jobs and their logs), \textit{data exploration} (query tables, check types), \textit{execution} (create branches, start a run)
    \item \texttt{smolagents}\footnote{https://huggingface.co/docs/smolagents/index} is the ReAct framework, handling the loop, tool calls and logs: \texttt{smolagents} performs reasoning in Python, making it effective at pipeline reasoning when compared to JSON-based tool calling   \cite{yang2024llmwizardcodewand,wang2024executablecodeactionselicit}\item LLM inference provided by OpenAI, Anthropic and TogetherAI through a configurable LiteLLM\footnote{https://docs.litellm.ai/} interface;
    \item a verifier function $Branch \rightarrow bool$, which is the ``proof-checking'' step before merging into main.
\end{itemize}

\begin{figure}
    \centering
    \includegraphics[width=\linewidth]{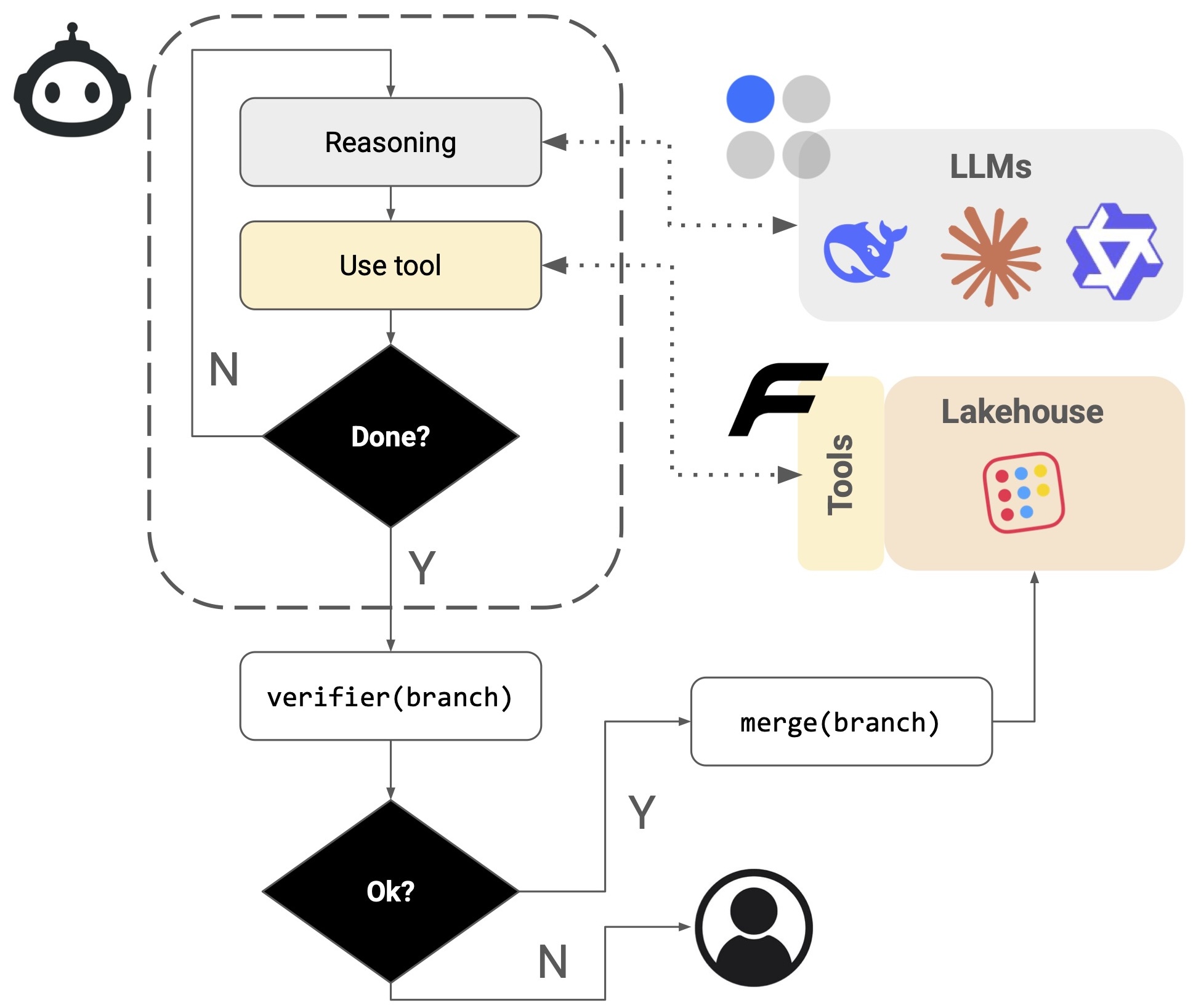}
    \caption{\textbf{Safe, untrusted lakehouse agents}: the agent leverages LLMs and tools to repair a data pipeline. When an answer is produced, a deterministic check in the ``outer loop'' verifies that it is safe to merge to production.}
    \label{fig:agent}
\end{figure}

\subsection{Running an experiment}

Fig.~\ref{fig:agent} illustrates the high-level flow of an experiment: the agentic loop (reasoning and tool usage), the underlying programmable lakehouse (entirely accessible through APIs, wrapped by the MCP server), the proof-checking step at the end. In the example scenario, the script first launches a faulty pipeline to create the conditions for agentic repair: following prior reports \cite{FOIDL2024111855} and industry experience, we simulate a package mismatch around the release of \texttt{NumPy 2.0} that caused crashes in containers using \texttt{pandas 2.0}.

A sample run\footnote{https://www.loom.com/share/fc79b52601074a5ba06e2f0272be3c62} using a model such as \textit{Sonnet 4.5} hits exactly all the abstractions discussed above: retrieving logs, querying the state of the lake, using declarative code to specify infrastructure changes, creating debug branches from production data, running code safely. Because this paper focuses on mapping abstractions for the agentic lakehouse and demonstrating feasibility, a thorough exploration of this experimental space is beyond scope. However, a few considerations are in order:

\begin{itemize}
    \item as a testament to the complexity of the task, frontier model performance vary greatly in success rate, token usage and number of tool calls; as system builders downstream from LLMs, the crucial takeaway is that even when models failed (e.g., \textit{GPT-5-mini}), the lakehouse exhibited no disruption or unsafe behavior;
    \item industry-leading traditional stacks -- such as Snowflake with dbt -- do not support agentic repair, even if they both have MCP servers and serve overlapping use cases. MCPs are a necessary but not sufficient condition for automation;
    \item because switching models is a single configuration change, we can easily imagine a budget-constrained scenario in which model selection is step-dependent, or a time-constrained situation in which data branches support concurrency control at scale for models in parallel.
\end{itemize}

\section{Conclusion and future work}

In a moment when most infrastructure agents are geared toward specific tasks \cite{gu2025argosagentictimeseriesanomaly}, programmable lakehouses have the ambition to support agentic reasoning across the full life-cycle of data. To the best of our knowledge, we addressed for the first time the open-ended challenge of repairing cloud pipelines. We argued that a programmable lakehouse is already agentic, and that declarative DAGs and Git-like data management are ideally suited to support safe-by-design agentic usage. 

To move beyond the hype, it is necessary to build novel systems and share working implementations. However, we also recognize that the infrastructure underlying agentic abstractions may need to evolve accordingly. Although massive parallelism is outside the scope of our prototype, it is arguably the primary challenge for OLAP systems in the age of agentic data exploration \cite{liu2025supportingaioverlordsredesigning}. We look forward to continuing to share with the community our research journey toward a fully agentic lakehouse.


\bibliographystyle{IEEEtran}
\bibliography{sample}

\end{document}